\documentclass[manuscript,nonacm]{acmart} 

\setcopyright{none}

\renewcommand\footnotetextcopyrightpermission[1]{}
\renewcommand{\footnotetextauthorsaddresses}[1]{}

\usepackage{pgfplots}
\usepackage{tikz}
\usetikzlibrary{positioning}
\pgfplotsset{compat=1.18}
\usepackage{makecell}
\usepackage{tabularx}
\usepackage{array}
\usepackage{subcaption}
\usepackage{xcolor}
\usepackage{booktabs}
\usepackage{graphicx}
\newcommand{\rev}[1]{\textcolor{black}{#1}}
\AtBeginDocument{%
  }

\begin{document}

\title{\rev{Argumentation-Based Person-Tailored Storytelling for Older Adults: A Reflective Hybrid AI Architecture}}

\author{Jayalakshmi Baskar}
\email{jayalakshmi.baskar@umu.se}
\author{Vera C. Kaelin}
\email{vera.kaelin@ost.ch}
\author{Kaan Kilic}
\email{kaan.kilic@umu.se}
\author{Helena Lindgren}
\email{helena.lindgren@umu.se}
\affiliation{%
  \institution{Umeå University, Department of Computing Science}
  \city{Umeå}
  \country{Sweden}
}

\renewcommand{\shortauthors}{Baskar et al.}

\begin{abstract}
This work investigates whether 
knowledge-driven, and large language model (LLM)-based storytelling can support purposeful narrative interaction with a digital companion for older adults. To address known limitations of LLMs, including hallucinations and limited transparency, we present a reflective storytelling agent that integrates knowledge graphs, user modelling, and argumentation theory to guide narrative generation, and 
\rev{applies argument mining as a diagnostic layer for inspecting argument structure, while persona grounding is assessed separately against structured persona information.}

The presented study consists of \rev{three} 
phases: Phase I employed a participatory design methodology, involving 11 domain experts in a formative evaluation that informed iterative refinement. The resulting system generates narratives grounded in structured user models representing health-promoting activities and motivations. 
Phase II involved 55 older adults who completed a survey evaluating persona-based narratives prompted across the argumentation-based dialogue purposes inquiry, deliberation, and persuasion 
at two creativity levels. Participants were asked about perceived 
purpose, relevance, cultural relatability, future-use intention and perceived inconsistencies. 
\rev{Phase III evaluated persona grounding across a full-system condition, an argumentation-scheme ablation, and a standard LLM baseline, and separately evaluated the zero-shot argument miner against a researcher-agreed consensus gold standard.}

\rev{In Phase II,} 
\rev{across the 330 narrative-purpose assessments after merging the two inquiry conditions,} the intended dialogue purpose was recognised in 30.6\%, while 33.3\% were categorised as having no clear purpose and 36.1\% as expressing another purpose. 
Cultural relatability showed a descriptive relationship with future-use intention, although more than half of participants reported that they would not use similar functionality in the future.

\rev{In Phase III, 67.6\% of applicable sampled statements in the full system were grounded in persona information, compared with 57.9\% without scheme guidance; this difference was not statistically significant. Scheme guidance nevertheless substantially changed argumentative organisation: narratives contained more explicit premises while claim counts remained essentially unchanged, whereas the full structured condition showed substantially greater sampled persona grounding than the standard baseline.} 
\rev{The zero-shot argument miner showed moderate performance compared to 
human-consensus annotation, supporting its use as an inspection aid rather than an autonomous verifier. Overall, the results demonstrate both the potential and limitations of 
reflective, person-tailored LLM storytelling for older adults. Our code, prompts, evaluation scripts, and reproducibility materials are publicly available at \url{https://github.com/jayalakshmi2k/reflective-storytelling-tist}}
\end{abstract}

\maketitle

\section{Introduction}\label{sec_introduction}

Storytelling is a fundamental mode of human expression through which people share experiences, convey values, and reflect on personally meaningful matters such as health and everyday activities \cite{bruner1991narrative}. Recent advances in Artificial Intelligence (AI), particularly LLMs, enable adaptive narratives tailored to individual users \cite{simon_tattletale_2022}. For older adults, personalised stories grounded in daily routines, motivations, and challenges may support emotional well-being and engagement in health-promoting activities. However, LLMs remain probabilistic and opaque, and can generate \textit{hallucinations}-outputs that are plausible but factually incorrect, logically inconsistent, or misaligned with user-provided information \cite{li2024survey,hicks_chatgpt_2024}. In health-related narratives, insufficient grounding risks misleading or inappropriate content.


Hybrid AI approaches that combine LLMs with symbolic knowledge representation and reflective mechanisms can reduce these risks \cite{Leofante24,steels2020personal}. Formal argumentation frameworks represent claims, premises, uncertainty, and contradiction, supporting inspection of reasoning in human-agent interaction \cite{baroni2018handbook,bench2003persuasion}. Argumentation schemes capture recurring reasoning patterns for purposes such as deliberation and persuasion \cite{walton2008argumentation}, and have been formalised in the Argument Interchange Format (AIF) for computational modelling \cite{Chesnevaretal06}. Prior work combining knowledge graphs with AIF supports personalised reasoning about health-related user information, including support for older adults \cite{YanLindgrenNieves18,lindgren2015acktus,baskar2015human,baskar2017multipurpose}. These approaches often build on standardised health classifications such as the World Health Organization (WHO)’s International Classification of Functioning, Disability and Health (ICF)\footnote{https://www.who.int/standards/classifications/international-classification-of-functioning-disability-and-health}, embedded in ontologies such as ACKTUS \cite{lindgren2015acktus}.

This research investigates how 
purposeful and personalised storytelling for older adults can be achieved through a hybrid AI architecture integrating knowledge graphs, user modelling, argumentation schemes, LLM-based narrative generation, and reflective inspection through argument mining.


Following a participatory design and knowledge engineering methodology \cite{Lindgrenetal2021}, we engage domain experts and older adults to study how narratives are perceived and evaluated. We introduce a \textit{Reflective Storytelling Agent} that constructs a knowledge-graph-based user model, interprets dialogue goals, selects a narrative strategy, and generates stories with an explicit communicative purpose. \rev{The agent then analyses its output through argument mining to identify argumentative units for reflective inspection; persona grounding is evaluated separately against the structured persona information.} We address the following research questions in 
\rev{three} complementary studies: 

\begin{enumerate}
  \item[\textbf{RQ1:}]  How do domain experts and older adults experience hybrid AI-generated personalised narratives in terms of purpose, cultural relatability, relevance to daily activities, and intention for future use?
  \item[\textbf{RQ2:}] To what extent do domain experts and older adults recognise argument-based narrative purposes and \textit{experience} levels of creativity in LLM-generated stories?
   \item[\textbf{RQ3:}] \rev{To what extent does argumentation-scheme guidance affect persona grounding and explicit argumentative structure, and how does the full structured architecture compare with a standard LLM baseline on persona grounding?}

\end{enumerate}

In this paper, we make the following contributions:
\begin{enumerate}
    \item We present a reflective personalised storytelling framework integrating ontology-based user modelling, argumentation schemes, LLM-based narrative generation, and reflective analytical layer for purpose-driven narratives for older adults.
    \item \rev{We introduce an argument-mining-based reflective layer that identifies argumentative units and structural patterns for reflective inspection.}
    \item We report a 
    \rev{three}-phase evaluation combining participatory design with domain experts and a 
    study with older adults, examining perceived purpose and experience \rev{together with a separate technical evaluation of persona grounding and argument-mining reliability.}
    \item \rev{We provide a controlled three-condition generation evaluation comparing the full architecture, an argumentation-scheme ablation, and a standard LLM baseline.}
    \item We derive design implications for reflective narrative systems that balance personalisation, purpose, and reliability in LLM-based storytelling.
\end{enumerate}

The remainder of this paper is structured as follows. Section~\ref{sec_background} reviews related work. Section~\ref{sec_methodology} presents the methodology for the 
\rev{three study phases.} 
Section~\ref{sec_system_overview} describes the system. Section~\ref{sec_results} reports results, followed by discussion (Section~\ref{sec_discussion}) and conclusions (Section~\ref{sec_conclusion}).

\section{Background and Related Work} \label{sec_background}


Digital storytelling for older adults has supported memory recollection, activity coaching, health feedback, reflection, and social connectedness \cite{rios_rincon_digital_2022,baskar2025towards,chang_digital_2023,riva2014positive,sehrawat_digital_2017}. Tailoring interaction to users' contexts, motives, and goals can support engagement \cite{van_velsen_tailoring_2019,baskar2015human}, but effectiveness also depends on how relevant content is selected and structured \cite{beinema2023discuss}. Limited empirical work has examined person-tailored storytelling that connects activities, motivations, and goals with concrete future-oriented scenarios \cite{rios_rincon_digital_2022,hinyard2007using,ten2024clarifying}.


LLMs support flexible narrative generation and personalisation based on user profiles, activities, goals, and communicative purposes \cite{openai_gpt5, wang_cue-cot_2023, liu_survey_2025, lucie_2-25_on-the-way}, but may produce content inconsistent with user information \cite{anh-hoang_survey_2025,asgari_framework_2025}. This is particularly problematic in health-related contexts \cite{hicks_chatgpt_2024}, motivating hybrid approaches combining generation with explicit constraints and knowledge structures \cite{li2024survey}.


Knowledge graphs and formal ontologies support structured reasoning, consistency, and explainability by explicitly representing entities, relations, and shared conceptual structures \cite{gruber1995toward}. In computational storytelling, they have been used to model narrative elements such as events, goals, and temporal relations, enabling purpose-aligned narrative generation \cite{finlayson2010computational}. Knowledge-driven approaches complement generative models by grounding narrative generation in structured representations, which is particularly important in health-related narratives where content must align with user data, domain knowledge, and ethical constraints \cite{besold2021neural, rios_rincon_digital_2022}. Ontologies have also been used to model argumentative structures and dialogue purposes, enabling systems to reason about narrative intent \cite{Chesnevaretal06}. 


Argumentation theory provides a framework for understanding how humans reason, justify claims, and engage in purposeful dialogue \cite{walton1995dialogues,walton2009argumentation}. Walton and Krabbe distinguish dialogue types based on communicative goals, such as \textit{persuasion}, \textit{deliberation}, and \textit{inquiry} \cite{walton1995dialogues}. In this work, these dialogue types are used to embed distinct narrative purposes in the agent to support health behaviour change grounded in personal values and priorities. Computational argumentation research has enabled these theoretical concepts to be operationalised in intelligent systems \cite{Dung95,BlackH09,bench2003persuasion,vanGijzel15argument}. Prior work in argumentation-based coaching and explanation systems shows that structured reasoning can improve transparency and user understanding, particularly in health-related contexts \cite{Hunter24,LindgrenKilic25,demollin_2020_argument,kilic2023argument}. Such systems can also model conflicting motives or values, for example tensions between health goals and limited time or motivation \cite{baskar2017multipurpose,Hunter24,Lindgrenetal25}. Argument quality influences the persuasive impact of narratives, especially when arguments are embedded within engaging story structures \cite{schreiner2018argument}. \textit{Argumentation schemes} capture recurring patterns of everyday reasoning by specifying premises, conclusions, and associated critical questions \cite{walton2008argumentation}. These schemes can support both the construction and assessment of argumentative content, enabling evaluation of whether narrative recommendations are well grounded \cite{walton2008argumentation,macagno2017argumentation}. Formal representations of argument schemes have been proposed to support reuse and sharing \cite{Chesnevaretal06}, and parts of this work have been implemented in the ACKTUS platform for knowledge-based health applications \cite{lindgren2015acktus}.

Human-AI collaboration benefits from transparency and user verification \cite{amershiHAI19, bansal21hat, Iftikhar24}. Reflective mechanisms including argument mining, constraint checking, and argument-based evaluation can surface potential issues for inspection in generated narratives \cite{guida_llms_2025,gordon_representing_2018,ruiz2025computational}.
\rev{Our earlier workshop paper \cite{baskar2025towards} introduced the person-tailored storytelling and collaborative-planning direction; the present work extends it through the reflective hybrid architecture, the expert and older-adult evaluations, and the controlled grounding, ablation, baseline, and argument-mining analyses.}


\section{Methodology} \label{sec_methodology}
\begin{figure}
    \centering
    \includegraphics[width=\linewidth]{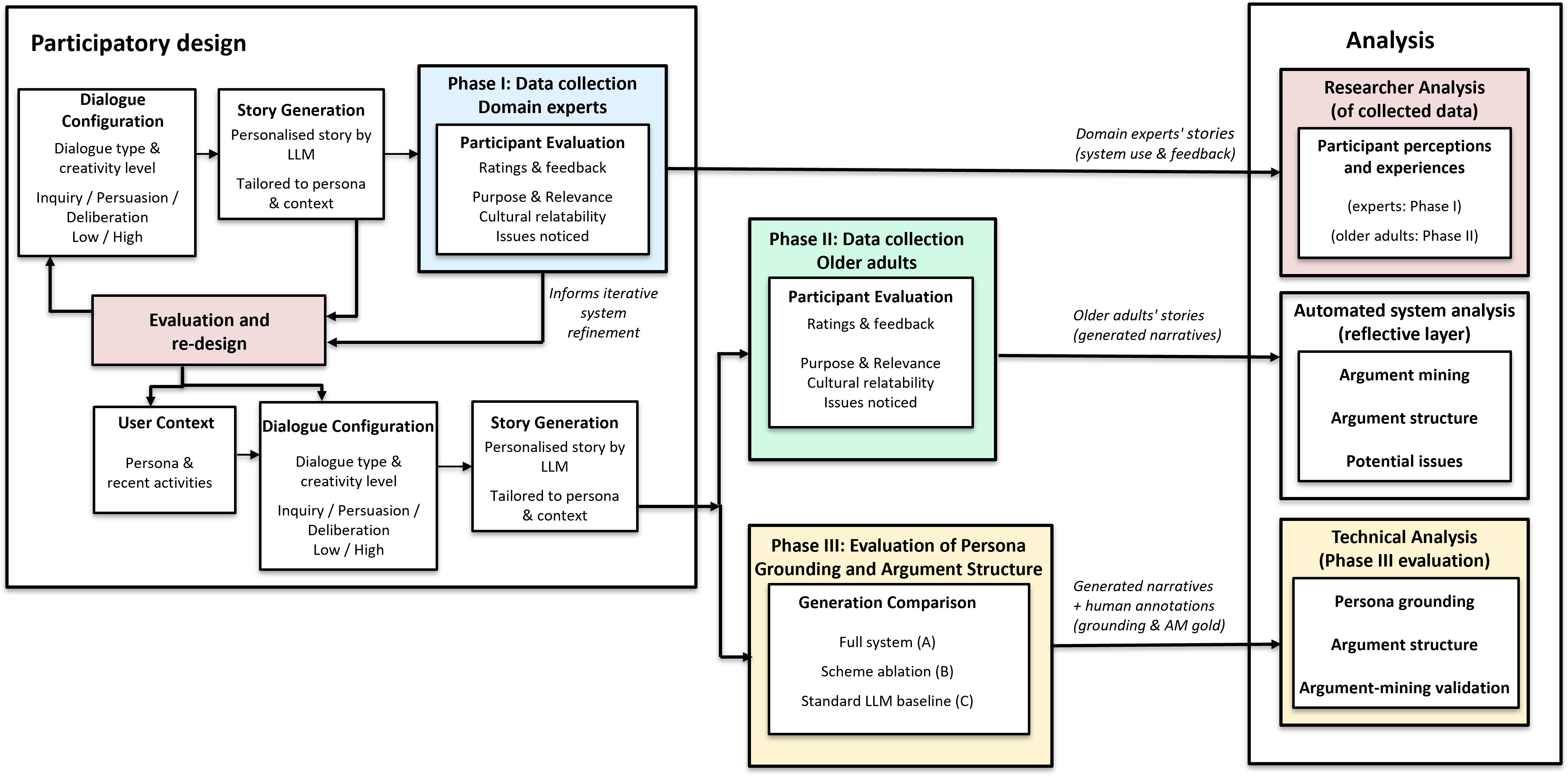}
   \caption{\rev{Overview of the three-phase study flow and system's reflective storytelling mechanisms. Phase I informed iterative system refinement, Phase II evaluated persona-based narratives with older adults, and Phase III evaluated persona grounding, argument structure, and argument-mining performance.}}
   \Description{}
    \label{fig:Study}
\end{figure}

\rev{The system and study were developed through a participatory design process involving a multidisciplinary research group with expertise in Occupational Therapy, Medical Informatics, Human-Computer Interaction, and Artificial Intelligence. Scenario-based methods informed the information model and narrative purposes \cite{baskar2025towards}. Dialogue types, argumentation schemes, and prompts were iteratively discussed, implemented, and trialled. The resulting study had three phases (Figure~\ref{fig:Study}). Phase~I was a formative evaluation with domain experts. Phase~II examined older adults' perceptions of persona-based narratives. Phase III evaluated persona grounding across three matched generation conditions: the full system (A), a scheme ablation (B), and a standard LLM baseline (C), and separately evaluated the zero-shot argument miner against a researcher-agreed consensus gold standard for Conditions A and B.}

\rev{In Phase I, domain experts participated in individual 60-minute online formative co-design sessions with an interviewer and note-taker. They entered activities from their lives, evaluated eight narratives spanning four dialogue prompts and two creativity levels, and completed a semi-structured interview on purpose, creativity, motivational value, relevance, risks, and system improvements.}

\subsection{Participants}\label{Part}

\rev{Phase I used convenience sampling of English-speaking healthcare professionals experienced in older adults and/or health promotion. Eleven experts (1 male, 10 female) from Sweden, Austria, and Switzerland participated: Occupational Therapy (n=9), Physiotherapy (n=1), and Nursing (n=2); professional-area counts were non-exclusive. }
\rev{The findings from Phase~I informed iterative refinements to the storytelling agent and study protocol before the Phase~II evaluation.}

In Phase~II, older adults were recruited via Prolific 
\rev{(\url{https://www.prolific.com/})},
which is an online platform for researchers to organise studies and gather data from a set of predefined subset of anonymous participants. The inclusion criteria defined were English fluency, age between 66 and 100 years, and be living in one of ten specified European countries. We compensated participants with 9€. Each session took on average 36 minutes. 
A total of 55 older adults (25 female, 30 male), aged 66–83 years (mean = 71), participated. Participants were primarily from the United Kingdom (n=30), followed by Portugal (n=5), Spain (n=5), the Netherlands (n=5), Germany (n=3), Sweden (n=2), France (n=2), Norway (n=1), Denmark (n=1), and Switzerland (n=1). The sample was concentrated in the younger age groups (65-69: 45.5\%; 70-75: 40.0\%; 76-85: 14.5\%).
\rev{Phase~III did not involve additional participants. It used narratives generated for five predefined personas and annotations produced by the research team.}



\subsection{Study Procedure and Materials: Personas, Activities, and Story Sets} \label{subsec_materials}

\rev{The study operationalised inquiry, deliberation, and persuasion through four prompts \cite{walton1995dialogues}. Narratives followed a beginning-middle-end structure, and participants evaluated eight stories. In Phase I, experts provided their own activities. In Phase II, older adults selected the most similar of five fictional personas (A-E), defined by structured activity and motivation profiles grounded in the activity ontology \cite{lindgren2022contextualising}. The personas represented physical, social, and recovery activities and included mobility, service-access, and social-network variation while avoiding gender stereotypes. Combining five personas with four dialogue prompts at low and high creativity produced 40 narratives.}

To examine the experience of creativity in AI-generated narratives, each story was generated in two versions: low and high creativity. \rev{Creativity was implemented through prompt-based style constraints, while keeping the dialogue type, argumentation scheme, user model, and narrative structure constant. Low-creativity stories used clear and simple language, whereas high-creativity stories allowed more expressive phrasing. No temperature-based manipulation was applied in the generation runs analysed in this study.}

\begingroup
\color{black}
Phase~III compared three matched generation conditions (Table~\ref{tab:generation_conditions}). Phase~III used a controlled ablation to examine the contribution of argumentation-scheme guidance. Condition~A (full system) included structured persona and activity information, preferences and priorities, grounding and safety instructions, dialogue-purpose guidance, and argumentation-scheme guidance. Condition~B contained the same information and instructions, except that the argumentation-scheme guidance was removed. Condition C was intentionally not supplied with persona information; therefore, its grounding evaluation measures whether its person-specific statements are supported by the predefined persona used for the matched experimental cell, not whether the baseline was instructed to reproduce that persona. 

Each condition contained 40 matched narratives across five personas and eight topics. All three conditions were generated using \texttt{gpt-5-mini-2025-08-07} via the OpenAI API. Condition A reused the Phase II narratives. A-B isolates the incremental effect of scheme guidance, whereas A-C compares the structured person-tailored architecture as a package with standard LLM generation. The complete generation prompts, the a priori dialogue-type--argumentation-scheme mappings, the low- and high-creativity style instructions, and the structured profiles of Personas A--E are provided in the public reproducibility repository. 
\endgroup

\begin{table}[t]
\color{black}
\caption{\rev{Matched Phase III generation conditions. A check mark denotes included information or guidance.}}
\label{tab:generation_conditions}
\centering
\begin{tabular}{lccc}
\toprule
\textbf{Prompt component} &
\textbf{A: Full} &
\textbf{B: No scheme} &
\textbf{C: Standard} \\
\midrule
Persona/activity information  & \checkmark & \checkmark & -- \\
Grounding instruction         & \checkmark & \checkmark & -- \\
Dialogue-purpose guidance     & \checkmark & \checkmark & -- \\
Argumentation scheme          & \checkmark & --         & -- \\
Matched topic/style/format    & \checkmark & \checkmark & \checkmark \\
Matched GenAI model     & \checkmark & \checkmark & \checkmark \\
\bottomrule
\end{tabular}
\end{table}

\rev{All participants provided informed consent before participation.}

\subsection{Data Collection}
\label{subsec_data_collection}

In Phase I, domain experts interacted with the system in semi-structured interviews focusing on creativity, communicative purpose, and motivational value. Sessions were recorded and transcribed. \rev{The qualitative feedback was used to iteratively refine the storytelling agent, prompting strategy, and study protocol before Phase~II.}

In Phase~II, participants completed background questionnaires, selected a persona, and evaluated each narrative with respect to perceived creativity, purpose recognition, motivational value, cultural relatability, perceived inconsistencies, and persona alignment, with optional explanations. Post-study items assessed overall experience, perceived relevance, and future-use intention. Attention checks were included to support data quality.

\subsection{Participant Evaluation Measures}

Participant evaluation measures in Phase~II were organised into 
(1) per-narrative measures collected after each \rev{generated} story and 
(2) post-study measures collected after 
\rev{participants had completed the full evaluation session }(Table~\ref{tab:measures}).
\rev{Table~\ref{tab:measures} summarises the Phase II evaluation measures.}

\paragraph{Per-narrative measures.}
After each narrative, participants evaluated purpose recognition, narrative appreciation, perceived creativity, and perceived inconsistency.

Purpose recognition was operationalised through predefined statements corresponding to inquiry, deliberation, and persuasion (e.g., highlighting past activities, inspiring future action, or providing reasons for change). Multiple selections were allowed. Recognition of the intended dialogue purpose was defined as the proportion of narratives for which participants selected at least one statement aligned with the predefined purpose.


\rev{Narrative appreciation and creativity were assessed on 5-point scales; creativity was additionally categorised as appealing, appropriate, too much, or too little.}

Perceived inconsistency captured whether participants noticed incorrect, implausible, or persona-inconsistent elements (yes/no/not sure), with optional explanation and disturbance rating. \rev{Disturbance ratings are summarised using the median and interquartile range (IQR) in Section~\ref{subsec_results_creativity}.}



\paragraph{\rev{Post-study measures.}} \rev{Participants reported overall experience, relevance, cultural relatability, and future-use intention (Table~\ref{tab:measures}). The complete questionnaire wording and the mapping of the purpose-response statements to inquiry, deliberation, and persuasion are provided with the study materials in the public reproducibility repository. }


\begin{table}[t]
\caption{Implementation of participant evaluation measures (Phase II: older-adult participants)}
\label{tab:measures}
\centering
\begin{tabular}{p{3.1cm} p{4.0cm} p{3.8cm} p{2.7cm}}
\toprule
\textbf{Construct} & \textbf{Operational definition} & \textbf{Measurement instrument} & \textbf{Analysis type} \\
\midrule

\multicolumn{4}{l}{\textit{Per-narrative evaluation measures}} \\
\midrule

Purpose recognition 
& Identification of perceived gain aligned with inquiry, deliberation, or persuasion 
& Multiple- choice checkboxes (purpose statements mapped to dialogue types) 
& Proportions; mapping to intended purpose \\

Narrative appreciation 
& Affective response to an individual narrative 
& 5-point Likert scale (dislike-like) 
& Descriptive statistics \\

Perceived creativity 
& Perceived creativity level of the narrative 
& 5-point Likert scale + categorical judgement (appealing, appropriate, too much, too little) 
& Descriptive statistics \\

Perceived inconsistency (issues) 
& Detection of incorrect or persona-inconsistent elements 
& Yes / No / Not sure + optional explanation + disturbance rating (5-point scale) 
& Proportions; \rev{disturbance-rating summary} \\

\midrule
\multicolumn{4}{l}{\textit{Post-study (overall) evaluation measures}} \\
\midrule

Overall experience 
& Holistic experience after reading all narratives 
& Open-ended reflection 
& Qualitative inspection \\

Overall relevance 
& Perceived personal relevance of narratives as a whole 
& 5-point Likert scale (low-high relevance) 
& Descriptive statistics \\

Cultural relatability 
& Perceived cultural fit with life context and values 
& Yes / Maybe / No + optional comments 
& Proportions; qualitative inspection \\

Future use intention 
& Willingness to use similar functionality in an app 
& Yes / Maybe / No + optional description 
& Proportions; qualitative inspection \\


\bottomrule
\end{tabular}
\end{table}


\subsection{Data Analysis} \label{subsec_methodology_data_analysis}


\rev{Phase~I used qualitative design-oriented analysis to identify recurring concerns, design recommendations, and refinements to the storytelling system.}

\rev{Phase~II used descriptive statistics to summarise responses to closed-ended questionnaire items, including frequencies and proportions. Responses to open-ended questions were analysed qualitatively to identify recurring themes, illustrative examples, and participants' interpretations of the generated narratives. For the low–high creativity comparison in Phase II, participant-level mean perceived-creativity and narrative-appreciation scores were compared using two-sided paired Wilcoxon signed-rank tests.}

\rev{Phase~III addressed RQ3 through a computational evaluation of argument mining performance and persona grounding. The evaluation consisted of two complementary analyses. First, the zero-shot LLM argument miner was evaluated against a researcher-agreed consensus gold standard using precision, recall, and F1-score for the three annotation labels (CLAIM, PREMISE, and NONE). Persona grounding was assessed primarily between Conditions A and B, for which the same persona information was available to the generator. Condition C was additionally coded using the same grounding framework to characterise the standard LLM baseline; because this condition did not receive persona information, A-C differences were interpreted as differences between the structured person-tailored architecture and the standard generation setup, rather than as an isolated test of grounding or argumentation-scheme effects. Statements were categorised as grounded, unsupported, or contradicted according to the corresponding persona activities, preferences, priorities, goals, and experiences. The combined analyses were used to examine whether argumentation-scheme guidance improved the explicit argumentative structure of the generated narratives while maintaining consistency with the predefined persona information. 
Persona grounding and argumentative support were treated as complementary but distinct properties. Persona grounding concerns whether a person-specific statement is supported by information in the external structured persona model, irrespective of whether the narrative explicitly states a reason for that statement. Argumentative structure concerns whether reasons are made explicit within the narrative itself through \textsc{Premise} units supporting claims. Consequently, a claim may be persona-grounded even when no explicit premise is expressed in the narrative. Conversely, the presence of a premise does not by itself establish that the premise is grounded in the persona information.}

\begingroup
\color{black}

\paragraph{Grounding annotation.}
A sample of 120 statements was selected from narratives generated under Conditions A, B, and C, with 40 statements sampled from each condition. Three researchers independently assessed the sampled statements for consistency with the corresponding predefined persona information. Each statement was labelled \textsc{Grounded} when person-specific content was supported by the persona information, \textsc{Unsupported} when it introduced person-specific content not supported by that information, \textsc{Contradicted} when it conflicted with the persona information, \textsc{Not Applicable} when it contained no person-specific content that could be assessed against the persona information, or \textsc{Unclear} when the available information or interpretation was insufficient for a confident judgement. Final reference labels were determined by majority vote among the three independent annotations. A majority label was obtained for 116 of the 120 statements; the four statements without a majority label were resolved through discussion among the annotators. The persona information used as evidence for a grounding judgement was an external annotation reference and should not be confused with a \textsc{Premise} expressed within the generated narrative.

\endgroup

\begingroup
\color{black}

\subsection{Consensus Gold Standard and Argument-Mining Evaluation}
\label{sec:am-method}

\paragraph{ADU segmentation and annotation.}
The narratives were initially segmented automatically into sentence-like units. Line breaks were replaced with spaces, after which each narrative was split at occurrences of a period followed by a space. Empty units were discarded, and the remaining units were numbered sequentially within each narrative. No linguistic sentence tokenizer was used.

The automatically produced units were then manually reviewed. When a unit contained a claim and a separable supporting justification, it was divided
into two argumentative discourse units (ADUs) so that the argumentative functions could be annotated separately. Each final ADU received one of three labels: \textsc{Claim}, \textsc{Premise}, or \textsc{None}. A \textsc{Claim} represented a recommendation, conclusion, interpretation, judgement, proposal, position, or other main argumentative point. A \textsc{Premise} provided a reason, justification, explanation, or item of evidence supporting a claim. \textsc{None} was assigned to narrative description, scene-setting, background information, greetings, or other content without a clear argumentative function.

Sentences containing clearly separable claims and justifications were divided during manual review. However, some units remained difficult to assign to a
single argumentative category because their functions were closely intertwined. One consensus label was retained for each such unit. These boundary cases are examined in the error analysis in Section~\ref{sec:error-analysis}.

\paragraph{Human-consensus gold standard.}
Two researchers independently segmented each generated narrative into argumentative discourse units (ADUs) and labelled each unit as CLAIM, PREMISE, or NONE. The two annotation sets were compared to identify agreements and disagreements. 
Agreements between the first two researchers were retained. Their disagreements were subsequently reviewed by two additional researchers. The resulting decisions were consolidated into the final human-consensus gold standard. The final evaluation files contained a consensus label for every evaluated ADU; no unlabeled or unresolved cases were included in the argument-mining evaluation.

The gold standard contained 529 ADUs from Dataset A and 436 ADUs from Dataset B, giving 965 evaluated ADUs in total. Table~\ref{tab:gold-composition} reports the final class distribution.

\begin{table}[t]
\color{black}
  \caption{\rev{Composition of the final human-consensus gold standard.}}
  \label{tab:gold-composition}
  \centering
  \small
  \begin{tabular}{lrrrr}
    \toprule
    Dataset & \textsc{Claim} & \textsc{Premise} & \textsc{None}
            & Total \\
    \midrule
    A: Full system     & 200 & 245 &  84 & 529 \\
    B: Scheme ablation & 201 &  49 & 186 & 436 \\
    \midrule
    Total              & 401 & 294 & 270 & 965 \\
    \bottomrule
  \end{tabular}
\end{table}

\paragraph{Zero-shot argument-mining evaluation.}
\texttt{gpt-5-mini-2025-08-07} was evaluated as a zero-shot argument miner against the frozen human-consensus gold standard. The model was not trained or fine-tuned using the annotated narratives. It received the annotation instructions, operational definitions of \textsc{Claim}, \textsc{Premise}, and \textsc{None}, and the narrative containing the target ADU.

For the primary evaluation, the complete narrative was provided as a sequence of indexed ADUs, with the target ADU explicitly identified. The model was
instructed to classify only the target ADU while using the surrounding narrative as context. Full-narrative context was used because the function of
an ADU, particularly its role as a premise, may depend on its relationship with neighbouring recommendations or conclusions.

The human-consensus label was not included in the model input. Gold labels were joined with the model predictions only after inference for the calculation of evaluation measures. The model output was restricted to one of the three permitted labels. A separate ADU-only evaluation, in which the target unit was classified without the complete narrative, was used as a sensitivity analysis to examine the contribution of narrative context.

Macro-averaged F1 was the primary evaluation measure because it assigns equal importance to \textsc{Claim}, \textsc{Premise}, and \textsc{None}, irrespective of differences in class frequency. Per-class precision, recall, F1-score, and support were also calculated. Accuracy was treated as a secondary measure, and confusion matrices were examined to identify class-specific error patterns. Results were reported separately for Datasets A and B because they represented different generation conditions and had substantially different class distributions.

A diagnostic error analysis complemented the quantitative evaluation. Full-corpus confusion counts were used to establish the frequencies of the different errors. 
A deterministic audit (seed 42) examined 45 cases: 30 errors sampled from the three dominant confusion types (12 Premise-to-Claim, 10 None-to-Claim, and 8 Premise-to-None) and 15 correct controls (five each for Claim, Premise, and None). The audit included cases from both Datasets A and B and exactly nine cases per persona. Each target ADU was examined together with neighbouring units and the complete narrative. The diagnostic review considered context dependence, ambiguous argumentative functions, possible gold-label inconsistencies, and clear model errors. These diagnostic judgements did not change the frozen consensus labels, and their frequencies were not treated as prevalence estimates for the complete dataset.

Argument-mining and persona-grounding evaluation addressed different questions. Argument-mining performance measured how accurately the model
identified argumentative functions relative to the human-consensus gold standard. Persona-grounding evaluation assessed whether generated statements were supported by the predefined persona's activities, preferences, priorities, goals, motivations, and experiences. Argument-mining performance was therefore not treated as evidence of persona grounding, factual correctness, hallucination detection, or general AI alignment.

\paragraph{Statistical analysis.}
All inferential tests were two-sided. For the sampled persona-grounding statements, condition-level proportions of \textsc{Grounded} versus \textsc{Unsupported} labels were compared using Fisher's exact tests based on applicable statements. Argument-structure outcomes were analysed using matched narratives. Paired narrative-level counts of \textsc{Claim}, \textsc{Premise}, \textsc{None}, and total argumentative units were compared using Wilcoxon signed-rank tests. The binary outcome indicating whether both a \textsc{Claim} and a \textsc{Premise} occurred within a narrative was compared between matched conditions using the exact two-sided McNemar test. Rank-biserial correlations were reported as effect sizes for the Wilcoxon tests. Confidence intervals for mean paired differences were estimated using 200,000 paired bootstrap resamples with seed 42, preserving the matched narrative pairs. Wilson 95\% confidence intervals were calculated for descriptive grounding proportions. Statistical significance was evaluated using $\alpha=.05$. Analyses were conducted using \texttt{Python 3.13.5}.
\endgroup





\section{Reflective Storytelling System: Framework and Implementation}
\label{sec_system_overview}

The personalised storytelling system is a hybrid narrative intelligence framework integrating ontology-based user modelling, formal argumentation, and LLM-based narrative generation. The system generates personalised, purpose-driven stories for older adults about their past and future planned activities, while enabling computational inspection of reasoning structure, narrative purpose, and potential grounding discrepancies. 

\subsection{Design Goals}
\label{subsec_design_goals}

The system was developed with four design goals:
\begin{enumerate}
    \item \textbf{Personalisation}: to generate narratives grounded in an ontology reflecting a user’s activities, motivations, goals, and lifestyle.
    \item \textbf{Formal argumentation}: to support purpose-driven narratives using argumentation schemes formalised through an ontology embedding a version of the AIF and embedded in LLM prompts.
    \item \textbf{\rev{Agent-based reflective inspection and transparency:}} \rev{to analyse generated narratives using an argument-mining module that identifies argumentative units and structural patterns for reflective inspection.}
    \item \textbf{\rev{Suitability for older adults:}} \rev{to provide a clear and accessible interaction flow for narrative presentation and feedback.}
\end{enumerate}

\subsection{System Architecture Overview}
\label{subsec_architecture}

\begin{figure*} [t]
    \centering
    \includegraphics[width=0.9\textwidth]{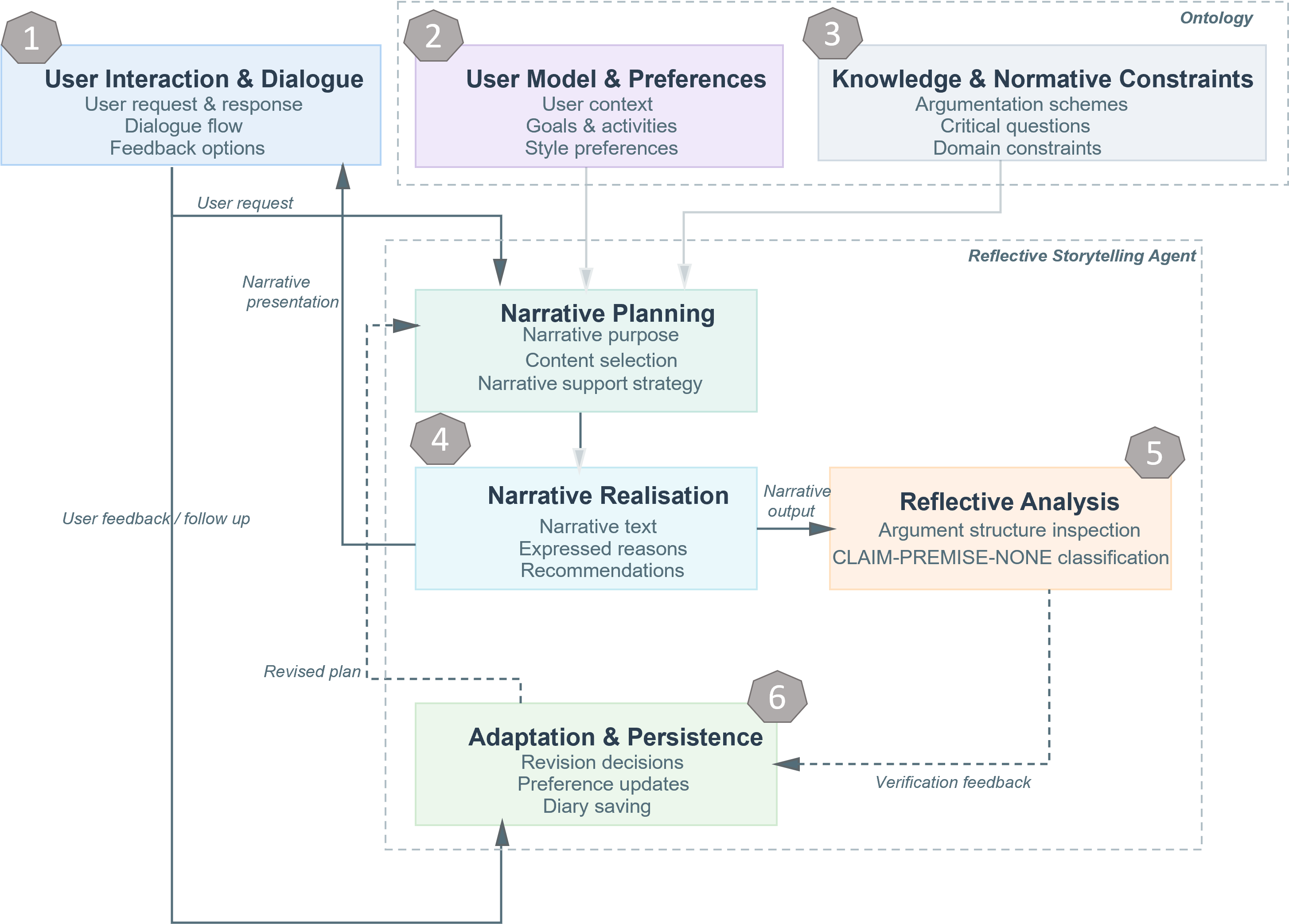}
    \caption{System architecture of the personalised reflective storytelling framework. \rev{The six levels of the system (numbered 1-6) integrates argumentation knowledge with LLM-based story generation, argument mining, user modelling, and interactive evaluation. Light-grey dashed boundaries indicate architectural groupings, while dotted arrows indicate prospective feedback or adaptation pathways that were not evaluated in the present study.}}
    \Description{A multi-layer architecture diagram showing the Knowledge Layer, User Modelling Layer, Story Generation Layer, Argument Mining Layer, Interactive Evaluation Layer, and SQL Data Storage. The knowledge layer lists four argumentation schemes and critical questions. Arrows show data flow between layers.}
    \label{fig:architecture}
\end{figure*}

The personalised storytelling system follows a layered architecture integrating narrative reasoning with generative models to support interpretable and purposeful story generation (Figure~\ref{fig:architecture}). The architecture comprises the following layers:

\begin{enumerate}
    \item \textbf{User interaction and dialogue}: supports user input, narrative presentation, and feedback collection through a web-based interface.
    \item \textbf{User model including preferences} (Section~\ref{subsec_usermodel}): represents activities, motivations, goals, and preferences structured into physical, social, and recovery domains.
    \item \textbf{Knowledge and argumentation constraints} (Section~\ref{subsec_storytelling_agent}): encodes domain knowledge, argumentation schemes, and critical questions that guide narrative purpose and reasoning structure.
    \item \textbf{Narrative generation}: comprises \emph{narrative planning}, which selects dialogue type, argumentation scheme, and content, and \emph{narrative realisation}, which generates the narrative text.
    \item \textbf{Reflective analysis} (Section~\ref{subsec_argument_mining}): \rev{analyses generated narratives through argument mining to identify claim and premise structure for reflective inspection.}
    \item \textbf{\rev{Prospective adaptation and persistence:}} \rev{represents how reflective-analysis outcomes and user feedback could inform future narrative generation; this pathway was not evaluated in the present study.}
\end{enumerate}

\subsection{Illustrative Walk-Through Example}
\label{subsec_walkthrough}

To illustrate how the components of the reflective storytelling framework \rev{are effectuated in the interaction between the agent and a person }in practice, 
we present the following scenario. 
An older adult requests a story about sustaining motivation for daily walking. 
The participant has earlier reported walking regularly, associated it with independence and mental well-being, and described it now as important but sometimes difficult to sustain. These inputs are represented in the user model as structured activity and value attributes.

Given this context, the agent selects an \textit{Argument from Value} scheme to support a persuasive purpose. The reasoning links the participant’s stated values (e.g., independence and well-being) to continued walking, suggesting that maintaining the activity promotes what the participant considers important.

The resulting argument plan, together with the user model, is passed to the LLM, which generates a personalised narrative illustrating how regular short walks support these values.

\subsection{The Ontology and User Model}
\label{subsec_usermodel}
In our system architecture, the knowledge layer is supported by ACKTUS, a Semantic Web platform embedding an ontology implemented as a knowledge graph using OWL\footnote{https://www.w3.org/OWL/} (the Web Ontology Language) and RDF\footnote{https://www.w3.org/RDF/} (the
Resource Description Framework) \cite{lindgren2015acktus}. The ontology represents user activities, motivations, and health-related concepts, as well as the core structures of argumentation dialogues. The reasoning layer builds on this knowledge representation through an adapted version of the Argument Interchange Format (AIF), which is used to represent argumentation schemes and their argument instances \cite{Chesnevaretal06, lindgren2015acktus}. The ontology is stored in an RDF4J\footnote{https://rdf4j.org/} repository and accessed via an API. In the implemented system, user activities and motivations retrieved through the interaction layer are grounded in the knowledge layer, while the reasoning layer uses the defined argumentation schemes to guide narrative generation.

The system maintains a user model aggregating activity-related information provided during logging, including activity type, motivational descriptions, social context (``done with'' relations), and numerical ratings of importance and fun. The user model serves as the primary input for narrative planning and reflective analysis. The model builds on the multipurpose motivational model~\cite{baskar2017multipurpose}, later operationalised in an activity ontology grounded in activity theory and behaviour change models~\cite{lindgren2022contextualising,lindgren2021conceptual}. 
Activities are logged across physical, social, and recovery domains and represented using structured attributes such as importance (1--5), fun (1--5), frequency, motivations, and social context. 
The user model functions as a shared representational layer, grounding narrative generation and \rev{providing the structured information against which persona grounding can be assessed separately.}

\subsection{Storytelling Agent: Argumentation-guided Narrative Generation}
\label{subsec_storytelling_agent}


\subsubsection{Narrative Planning}
\label{subsec_narrative_planning}
For each narrative request, the agent derives an \textit{argument plan} specifying dialogue type (Persuasion, Deliberation, or Inquiry)~\cite{walton1995dialogues}, selected argumentation scheme, grounded premises, and a scheme-oriented conclusion. Critical questions associated with the scheme are used to support reflective reasoning. The argument plan guides narrative structure, focus, and communicative purpose. Narratives are generated using a fixed three-paragraph structure and style constraints tailored to older adults. The agent currently draws on four argumentation schemes represented in ACKTUS: \textit{Argument from Value}, \textit{Argument from Position to Know}, \textit{Sufficient Condition Scheme}, and \textit{Argument from Expert Opinion}~\cite{walton2008argumentation,lindgren2015acktus}. In this study, persuasive narratives primarily use Argument from Value, deliberative narratives use the Sufficient Condition Scheme, and inquiry narratives use either Argument from Value or Argument from Position to Know. Argument from Expert Opinion is represented but not activated. \rev{This mapping was specified a priori rather than inferred from generated narratives. Argument from Value was selected for persuasion because it connects proposed actions to personally important values; Sufficient Condition was used for deliberation to structure reasoning about possible future actions and outcomes; and Position to Know was used for inquiry when the narrative reflected on information available from the person’s recorded activities. Argument from Expert Opinion was retained in the ontology but not used because the study narratives did not introduce an external expert source.}

\subsubsection{Narrative Realisation}
\label{subsec_narrative_realisation}

The narrative generation layer combines the user model with formal argumentation theory to guide narrative construction. 
The system implements creativity as an explicit control parameter. For each request, the storytelling agent passes a creativity level (low or high) to the language model together with the structured prompt, determining generation style constraints. 
The intended argument plan and persona information supplied to the model were held constant across creativity levels. Creativity manipulation therefore occurred through style instructions, although the resulting narratives could still differ in content or reasoning because generation is stochastic. Generated narratives are presented to participants and forwarded to the argument mining module.

\subsection{Argument Mining Module}
\label{subsec_argument_mining}

Argument mining aims to identify argumentative components (e.g., claims and premises) and their structural relations in text \cite{mochales2011argumentation,lippi2016argumentation}. Scheme-based approaches further analyse arguments relative to predefined reasoning patterns \cite{walton2008argumentation}. In our system, argument mining is implemented as a constrained, scheme-guided reflective analysis of LLM-generated narratives relative to a structured user model.


This reflective layer does not verify factual correctness or autonomously detect hallucinations. In the revised evaluation, its primary role is argument-structure inspection: identifying CLAIM, PREMISE, and NONE units for comparison with the human-consensus annotation. Persona grounding is evaluated separately through researcher coding rather than inferred from argument-mining performance.

\section{Results} \label{sec_results}
\rev{Section \ref{sec_DE} reports the formative expert findings in Phase I, and Section \ref{sec_OA} reports the older-adult evaluation in Phase II. Section \ref{sec:phase3} reports persona grounding across Conditions A, B, and C. Section \ref{sec:am-results} evaluates the zero-shot argument miner against the human-consensus gold standard. Section \ref{sec:statistical-comparison} reports matched statistical comparisons, and Section \ref{sec:error-analysis} presents representative error cases.}




\subsection{Phase I: Domain Experts' Perspectives}
\label{sec_DE}

\begingroup
\color{black}


\subsubsection{Grounding in Activities and Motivations (RQ1)}

Experts valued narratives that connected their chosen 
activities with their associated motivations and everyday contexts. References to physical and social activities could make a story feel relevant when they remained consistent with the information entered into the system.

Experts also identified narratives that extended beyond the available activity information. Reported examples included inferred emotional states, social preferences, or motives, such as a need for relaxation or a wish to deepen relationships, that had not been supplied by the expert. Some elaborations drew on stereotypical assumptions. For example, an interest in watching sport was expanded into a competitive, boundary-pushing orientation that did not fit the represented older female user. These examples were treated as unsupported or stereotypical inferences rather than as verified facts about the represented person. Experts expressed 
that such mismatches could reduce the perceived appropriateness and trustworthiness of a narrative.

\subsubsection{Creativity, Clarity, and Narrative Tone (RQ1 and RQ2)}

Experts perceived differences between the low- and high-creativity variants. Low-creativity narratives were described as clearer and easier to follow. High-creativity narratives could provide richer and more engaging expression, but some also contained vivid descriptions, complex language, or speculative elaborations that added little to the intended meaning or moved away from their chosen 
activities. Experts therefore raised concerns that highly expressive language could become cognitively demanding for some older adults.

Tone was also important. Some narratives framed everyday activities as obligations, particularly when the recorded motivation included expressions such as ``it has to be done'' or ``others expect me to do it.'' In these cases, activities such as walking a dog or having dinner could be presented as burdensome routines rather than as opportunities for enjoyment, companionship, rest, or personal choice. Experts observed that this obligation-driven wording could make a narrative sound negative or prescriptive.

\subsubsection{Recognition of Narrative Purpose (RQ2)}

Since Phase I did not report purpose-recognition counts or a quantitative accuracy analysis, the following findings are presented as qualitative observations rather than comparative recognition rates.

The results indicated that the persuasion and deliberation purposes were more readily associated with motivational reasoning, suggestions, and consideration of future action. Inquiry was less readily distinguished, particularly when a narrative reflected on motives or completed activities. Some inquiry narratives were instead interpreted as advice or as encouraging future action. These observations suggest that the intended inquiry framing was not always evident to the experts. 

\subsubsection{Implications for System and Study Design}
\label{Design}

The Phase I findings 
led to several changes in system design and study setup for 
Phase II. First, experts found some narratives lengthy and insufficiently connected to the supplied activity information. In response, the narratives were shortened, and the generation prompt was revised to emphasise specificity and consistency with the available activity and motivation information.

Second, experts observed that highly creative narratives could become elaborate, linguistically complex, or speculative. The high-creativity prompt was therefore constrained to reduce excessive stylistic embellishment and narrow the difference between the low- and high-creativity variants while retaining some expressive variation.

Third, experts identified obligation-driven and potentially prescriptive wording. Prompt instructions were revised to allow routine activities to be framed in relation to rest, connection, enjoyment, or personal choice when supported by the supplied motivation, rather than presenting them only as tasks that should be completed.

Fourth, comments concerning stereotypical assumptions informed greater attention to gender- and age-related stereotyping in the subsequent system and study design. Experts also noted that cultural context could influence how a narrative was interpreted. A cultural-relatability item was consequently included in Phase II, and participants were recruited from the United Kingdom and several other European countries to increase cultural diversity. 


\endgroup

\subsection{Phase II: Older Adults' Perspectives}
\label{sec_OA}

\begingroup
\color{black}

Phase II addressed RQ1 by examining older adults' experiences of the persona-based narratives in terms of relevance, cultural relatability, intention for future use, and recognition of narrative purpose. Participants did not receive narratives generated from their own activity information. Instead, each participant selected one of five predefined fictional personas that they considered relatively similar to themselves. The findings therefore concern the perceived relevance and cultural fit of persona-based narratives rather than real-user personalisation or objective grounding.

\subsubsection{Perceived Relevance}
\label{subsec_results_relevance}


\rev{Overall relevance was assessed once after participants had evaluated all eight narratives. The mean rating was 2.62 on the five-point scale (median = 2; range = 1--5), indicating substantial variation in how personally relevant participants found the narratives as a whole.} 

Usefulness was not measured through a separate direct questionnaire item. Consequently, relevance and future-use intention are reported as distinct outcomes and are not treated as direct measures of usefulness.

\subsubsection{Cultural Relatability and Future-Use Intention}
\label{subsec_results_cultural_future}

Cultural relatability and future-use intention were assessed once per participant after all eight narratives had been evaluated. Twenty-four of the 55 participants (43.6\%) answered that the narratives were culturally relatable, 17/55 (30.9\%) answered ``Maybe,'' and 14/55 (25.5\%) answered ``No.''

Only 9/55 participants (16.4\%) reported that they would use similar storytelling functionality in the future. A further 16/55 (29.1\%) answered ``Maybe,'' whereas 30/55 (54.5\%) answered ``No.'' Thus, more than half of the participants did not express an intention to use similar functionality in the future.


\begin{table}[t]
\caption{Future-use intention by perceived cultural relatability of the persona-based narratives (N=55).}
\label{tab:futureuse_cultural}
\centering
\begin{tabular}{lrrrr}
\toprule
& \multicolumn{3}{c}{\textbf{\rev{Cultural relatability}}} & \\
\cmidrule(lr){2-4}
\textbf{Future-use intention} &
\textbf{Yes} &
\textbf{Maybe} &
\textbf{No} &
\textbf{Total} \\
\midrule
Yes   & 8  & 1  & 0  & 9  \\
Maybe & 11 & 4  & 1  & 16 \\
No    & 5  & 12 & 13 & 30 \\
\midrule
\textbf{Total} & 24 & 17 & 14 & 55 \\
\bottomrule
\end{tabular}

\begin{minipage}{0.96\linewidth}
\footnotesize
Columns report responses to cultural relatability. Rows report future-use intention. The table presents a descriptive cross-tabulation; no inferential association is claimed.
\end{minipage}
\end{table}

Median persona relatedness was 5.0/5 among both participants answering ``Yes'' and those answering ``Maybe'' to future use, and 4.5/5 among participants answering ``No.'' The relatively high median in all three groups indicates that persona relatedness alone did not distinguish participants who intended to use the functionality from those who did not.

\subsubsection{Purpose Recognition}
\label{subsec_results_purpose}

After each narrative, participants could select multiple statements describing what they perceived that they gained from the narrative. These statements represented inquiry, deliberation, and persuasion purposes. Because participants could select multiple statements, purpose recognition should not be interpreted as unique classification of a narrative.

The study produced 466 individual purpose selections. For comparison across three dialogue types and two creativity levels, the purpose analysis was normalised to 330 assessments, corresponding to 55 participants across six dialogue-type--creativity conditions. The intended purpose was recognised in 101/330 assessments (30.6\%). No clear purpose was identified in 110/330 assessments (33.3\%), while 119/330 assessments (36.1\%) were categorised as containing another, non-intended purpose.


\begin{table}[t]
\caption{Assessment-level purpose recognition by dialogue type and creativity level. Each row contains 55 normalised assessments. Intended-purpose recognition indicates that the assessment was categorised as containing the intended dialogue purpose.}
\label{tab:app-purpose-recognition}
\centering
\begin{tabular}{llrrr}
\toprule
\textbf{Dialogue type} &
\textbf{Creativity} &
\makecell{\textbf{No clear}\\\textbf{purpose}} &
\makecell{\textbf{Intended}\\\textbf{purpose}} &
\makecell{\textbf{Other}\\\textbf{purpose}} \\
\midrule
Persuasion   & Low  & 31 & 11 & 13 \\
Persuasion   & High & 21 & 12 & 22 \\
Deliberation & Low  & 13 & 23 & 19 \\
Deliberation & High & 12 & 16 & 27 \\
Inquiry      & Low  & 15 & 22 & 18 \\
Inquiry      & High & 18 & 17 & 20 \\
\midrule
\multicolumn{2}{l}{\textbf{Total}} &
110/330 (33.3\%) &
101/330 (30.6\%) &
119/330 (36.1\%) \\
\bottomrule
\end{tabular}
\end{table}

\subsubsection{Perceived Creativity, Narrative Appreciation, and Inconsistencies}
\label{subsec_results_creativity}
The intended low--high creativity manipulation was not reflected in participants' ratings in the expected direction. Participant-averaged perceived-creativity ratings were slightly lower for high-creativity than low-creativity narratives (2.65 vs.\ 2.76; paired Wilcoxon $W=213$, $p=.033$). Narrative appreciation showed a similar descriptive direction (2.87 vs.\ 2.98), but the difference was not statistically significant ($W=295$, $p=.178$). Among 439 valid categorical creativity judgements, ``appropriate'' was the most frequent response (208/439, 47.4\%); ``too little'' creativity was selected in 126/439 (28.7\%) evaluations, compared with 49/439 (11.2\%) judged as ``too much''. Across the eight narratives, participants provided 440 narrative-level assessments of whether they noticed issues or inconsistencies relative to their selected persona. Issues were reported in 51 assessments (11.6\%), while 39 (8.9\%) were rated ``Not sure'' and 350 (79.5\%) were rated ``No.'' Among the 75 ``Yes'' or ``Not sure'' assessments with a valid bother rating, the median rating was 2 on the five-point scale (IQR 1--3; range 1--4). These responses represent participants' subjective perceptions and were not treated as verified instances of factual error or persona-grounding failure.

\endgroup

\begingroup
\color{black}

\subsection{Phase III: Persona Grounding and Argument Structure}
\label{sec:phase3}

For the sampled persona-grounding statements, condition-level proportions of \textsc{Grounded} versus \textsc{Unsupported} labels were compared using two-sided Fisher's exact tests on applicable statements. Separately, argument-structure outcomes for Conditions A and B were analysed at the matched narrative level. Paired narrative-level counts of \textsc{Claim}, \textsc{Premise}, \textsc{None}, and total argumentative units were compared using two-sided Wilcoxon signed-rank tests. The binary outcome indicating whether both a \textsc{Claim} and a \textsc{Premise} occurred within a narrative was compared between matched A--B narrative pairs using the exact two-sided McNemar test. Rank-biserial correlation was reported for Wilcoxon tests, and confidence intervals for mean paired differences were estimated using 200,000 paired bootstrap resamples with seed 42. Statistical significance was evaluated using $\alpha=.05$.

\subsubsection{Persona Grounding Across Generation Conditions}



The A--C comparison provides evidence that the structured generation architecture produced substantially more content demonstrably connected to the supplied persona information than the standard generation baseline. This comparison evaluates the structured architecture as a package: Condition A included persona and activity information, grounding instructions, dialogue-purpose guidance, and argumentation-scheme guidance, whereas Condition C did not. It therefore does not isolate the contribution of any single component. Moreover, because Condition C was not supplied with persona information, its zero grounding rate should not be interpreted as showing that a standard LLM is intrinsically unable to produce appropriate narratives.


Condition A had a descriptively higher applicable grounding rate than Condition B (67.6\% versus 57.9\%). However, the difference was not statistically significant (two-sided Fisher's exact test, $p=.468$). Thus, the sampled grounding analysis did not establish an independent improvement in persona grounding from argumentation-scheme guidance. Importantly, this grounding comparison assesses consistency with the external persona information; it does not assess whether supporting reasons were made explicit within the narrative.

\begin{table}[t]
  \caption{\rev{Persona grounding across the full system, scheme ablation, and
  standard LLM baseline. Rates and Wilson 95\% confidence intervals use
  applicable statements as the denominator. NA denotes statements for which
  persona grounding was not applicable. One Condition C statement was labelled Unclear and was excluded from the
applicable-statement denominator.}}
  \label{tab:grounding_conditions}
  \centering
  \small
  \rev{
  \begin{tabular}{lrrrrrr}
    \toprule
    Condition & Applicable & Grounded & Unsupported & Contradicted & NA &
    Grounded rate [95\% CI] \\
    \midrule
    A: Full system    & 34 & 23 & 11 & 0 & 6  & 67.6\% [50.8, 80.9] \\
    B: Without schemes & 38 & 22 & 16 & 0 & 2  & 57.9\% [42.2, 72.1] \\
    C: Standard LLM   & 28 & 0  & 28 & 0 & 11 & 0.0\% [0.0, 12.1] \\
    \bottomrule
  \end{tabular}
  }
\end{table}



\subsection{Argument Mining}
\label{sec:am-results}


Table~\ref{tab:am-performance} reports the context-aware results. Dataset A had a macro-F1 of .612, compared with .350 for Dataset B. This is a descriptive difference between datasets whose generated narrative structures and class distributions differ. It should not be interpreted as evidence that
argumentation-scheme guidance inherently improves the argument-mining model.

At the class level, \textsc{Claim} and \textsc{Premise} were identified more
successfully than \textsc{None}. This weakness indicates that descriptive, rhetorical, and other non-argumentative narrative units were frequently interpreted as
argumentative.

\begin{table}[t]
\color{black}
  \caption{\rev{Context-aware zero-shot argument-mining performance against the
  frozen human-consensus gold standard. P, R, and F1 denote precision, recall,
  and F1-score. Macro averages assign equal weight to \textsc{Claim},
  \textsc{Premise}, and \textsc{None}.}}
  \label{tab:am-performance}
  \centering
  \small
  \begin{tabular}{llrrrrr}
    \toprule
    Dataset & Class/Average & P & R & F1 & Support & Accuracy \\
    \midrule
    A & Macro average       & .640 & .628 & .612 & 529 & .698 \\
    A & \textsc{Claim}      & .676 & .990 & .803 & 200 & -- \\
    A & \textsc{Premise}    & .918 & .596 & .723 & 245 & -- \\
    A & \textsc{None}       & .325 & .298 & .311 &  84 & -- \\
    \midrule
    B & Macro average       & .452 & .432 & .350 & 436 & .454 \\
    B & \textsc{Claim}      & .467 & .851 & .603 & 201 & -- \\
    B & \textsc{Premise}    & .351 & .408 & .377 &  49 & -- \\
    B & \textsc{None}       & .538 & .038 & .070 & 186 & -- \\
    \bottomrule
  \end{tabular}
\end{table}

Figure~\ref{fig:am-confusion} shows the context-aware confusion matrices for Datasets A and B. In Dataset A, 198/200 gold \textsc{Claim} ADUs were correctly identified. However, 47/245 gold \textsc{Premise} ADUs and 48/84 gold \textsc{None} ADUs were classified as \textsc{Claim}. A further 52/245 premises were classified as \textsc{None}, showing that implicit or context-dependent supporting roles remained difficult to identify.

The most pronounced error in Dataset B was the classification of gold \textsc{None} ADUs as \textsc{Claim}: 168/186 such units, or 90.3\%, received this prediction. Dataset B also showed greater confusion between claims and premises: 26/201 gold claims were classified as premises, while 27/49 gold premises were classified as claims. Thus, the model tended to over-identify argumentative conclusions, particularly in narratives containing fewer
explicit premises.

\begin{figure}[t]
  \centering
  \includegraphics[width=\linewidth]{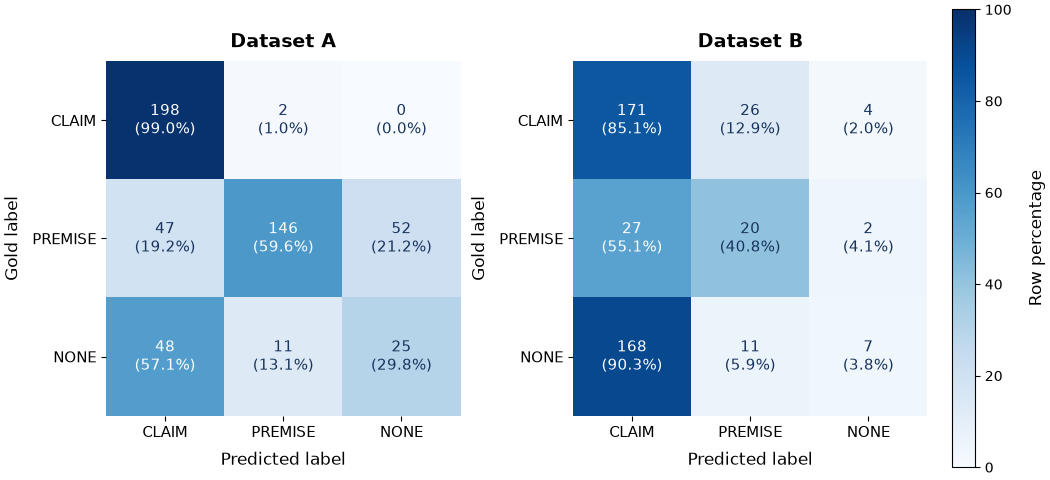}
  \caption{\rev{Context-aware argument-mining confusion matrices for
  (a) Dataset A, generated with argumentation-scheme guidance, and
  (b) Dataset B, generated without scheme guidance. Rows represent the human
  gold labels and columns represent the predicted labels. Cell percentages
  are normalised within each gold-label row; counts are shown separately for
  A and B.}}
  \label{fig:am-confusion}
  \Description{}
\end{figure}

As a secondary sensitivity analysis, the argument miner was also evaluated when each ADU was classified without its surrounding narrative. In this restricted-input setting, combined macro-F1 was .375 and accuracy was .498, compared with .508 and .588, respectively, when full narrative context was provided. The largest class-level difference concerned \textsc{Premise}, whose F1 increased from .113 with ADU-only input to .651 with narrative context. This observed improvement is consistent with the relational nature of premises: whether a statement supports a conclusion may only become clear from nearby statements. Nevertheless, narrative context did not resolve the weak recognition of \textsc{None}, and substantial classification errors remained.

\subsection{Statistical Comparison}
\label{sec:statistical-comparison}

Matched statistical comparisons were conducted for two distinct purposes. The A--B comparison assessed the incremental contribution of argumentation-scheme guidance because all other structured inputs and instructions were retained in both conditions. The A--C comparison assessed the complete structured architecture relative to the standard generation baseline. As described in Section~\ref{sec:phase3}, this latter comparison changes several prompt components simultaneously and therefore does not isolate the contribution of argumentation schemes or any other individual component.

\subsubsection{Full System Versus Scheme Ablation}


The scheme instruction was, however, associated with differences in explicit argument structure. Condition A contained more premises per narrative than
Condition B (means 6.13 and 1.23; mean paired difference 4.90, bootstrap 95\% CI [4.28, 5.53]). The Wilcoxon signed-rank test supported this difference
($W=0$, $p=3.24\times10^{-8}$; rank-biserial $r=1.000$). Condition A also contained more argumentative units per narrative (means 11.13 and 6.25; mean paired difference 4.88, bootstrap 95\% CI [4.03, 5.73]; $W=0$, $p=7.29\times10^{-8}$; rank-biserial $r=1.000$) and fewer \textsc{None} units (means 2.10 and 4.65; mean paired difference $-2.55$, bootstrap 95\% CI [$-3.33$, $-1.80$]; $W=21$, $p=3.23\times10^{-6}$; rank-biserial $r=-.925$).

Claims per narrative did not differ statistically between A and B (means 5.00 and 5.03; mean paired difference $-0.03$, bootstrap 95\% CI [$-0.85$, 0.78]; $W=238.5$, $p=.851$; rank-biserial $r=.038$). All 40 narratives in A contained both a \textsc{Claim} and a \textsc{Premise}, compared with 30/40 narratives in B. The exact two-sided McNemar test supported this difference (A-only = 10, B-only = 0; $p=.00195$), with a paired risk difference of 25.0 percentage points.

\begin{table}[t]
\color{black}
\caption{\rev{Matched statistical comparisons between generation conditions.
Mean differences are A minus B. Confidence intervals (CIs) were estimated
using 200,000 paired bootstrap resamples. RB denotes rank-biserial
correlation and RD denotes paired risk difference.}}
\label{tab:statistical-comparisons}
\centering
\small
\setlength{\tabcolsep}{4pt}
\renewcommand{\arraystretch}{1.12}

\begin{tabular}{@{}llccc@{}}
\toprule
\textbf{Comparison} &
\textbf{Outcome} &
\textbf{Effect [95\% CI]} &
\textbf{Test} &
\textbf{Effect size} \\
\midrule

A--B &
\textsc{Premises}/narrative &
$+4.90$ [4.28, 5.53] &
$W=0$, $p=3.24\times10^{-8}$ &
RB $=1.000$ \\

A--B &
\textsc{Claims}/narrative &
$-0.03$ [$-0.85$, 0.78] &
$W=238.5$, $p=.851$ &
RB $=.038$ \\

A--B &
\textsc{None}/narrative &
$-2.55$ [$-3.33$, $-1.80$] &
$W=21$, $p=3.23\times10^{-6}$ &
RB $=-.925$ \\

A--B &
Argumentative units/narrative &
$+4.88$ [4.03, 5.73] &
$W=0$, $p=7.29\times10^{-8}$ &
RB $=1.000$ \\

A--B &
Both \textsc{Claim} and \textsc{Premise} &
$+25.0$ pp &
McNemar, $p=.00195$ &
RD $=.250$ \\

\bottomrule
\end{tabular}
\end{table}

\subsection{Error Analysis}
\label{sec:error-analysis}


\subsubsection{Argument-Mining Successes and Failures}

As shown in Table~\ref{tab:error-examples}, the argument miner correctly identified explicit recommendations as \textsc{Claim} and supporting considerations as \textsc{Premise}. Narrative context was particularly helpful when the argumentative role depended on an adjacent recommendation. For example, ``Movement raises energy and keeps joints flexible'' was classified as \textsc{Claim} when presented alone but as \textsc{Premise} when it followed a recommendation to exercise. This change agreed with the consensus label. The model also correctly recognised some short descriptive or scene-setting units as \textsc{None}. Such successes were less common for \textsc{None} than for the two argumentative classes, consistent with the per-class results in Section~\ref{sec:am-results}.

The dominant full-corpus error was \textsc{None}-to-\textsc{Claim}, accounting for 216/398 errors (54.3\%), including 168 cases in Dataset B. Rhetorical encouragement was particularly difficult. For example, ``Keep up the great work, as each step you take adds meaning to your life and those around you'' was labelled \textsc{None} in the consensus standard but predicted as \textsc{Claim}, even with narrative context. The model therefore tended to interpret encouraging conclusions as substantive argumentative positions.

The second most frequent error was \textsc{Premise}-to-\textsc{Claim} (74/398; 18.6\%). Some cases reflected a genuinely relational role that could
only be established from the surrounding recommendation or conclusion. However, context did not resolve every case. The statement ``Group aerobics, caring for grandchildren, and your community work brought real connection and clear purpose'' remained classified as \textsc{Claim}, although its consensus label was \textsc{Premise} because it supported the subsequent conclusion that the persona had done something important.

A further 54/398 errors (13.6\%) were \textsc{Premise}-to-\textsc{None}. These cases often involved activity descriptions whose supporting function was implicit. Less frequent errors were \textsc{Claim}-to-\textsc{Premise} (28/398; 7.0\%), \textsc{None}-to-\textsc{Premise} (22/398; 5.5\%), and \textsc{Claim}-to-\textsc{None} (4/398; 1.0\%).

The diagnostic audit also exposed limitations of assigning exactly one label to every ADU. For example, ``You could increase your cycling trips since it
combines physical exercise with a bit of fun'' contains both a recommendation and its justification. Its consensus label was \textsc{Premise}, while the
model predicted \textsc{Claim} with and without context. The recommendation supports the model prediction, whereas the causal justification supports the consensus label. The consensus label was retained as the evaluation reference, but the case illustrates ambiguity introduced by mixed ADUs.

Within the 30 deliberately selected errors, 13 were diagnosed as dependent on missing context, 12 as possible residual gold-label inconsistencies, three as
mixed or ambiguous ADUs, and two as clear model errors. These proportions do not describe the complete error corpus because the audit intentionally over-sampled analytically informative errors. They indicate that observed disagreement arose from both model limitations and difficult boundaries in the single-label annotation task.

\begin{table}[t]
\color{black}
  \caption{\rev{Selected cases illustrating the main argument-mining and
  persona-grounding patterns. AM labels are reported only for Conditions A and
  B. Grounding labels are researcher judgements against the predefined persona
  information and are not AM predictions.}}
  \label{tab:error-examples}
  \centering
  \small
    \begin{tabularx}{\linewidth}{
    p{1.45cm}
    p{0.50cm}
    X
    p{1.60cm}
    p{1.85cm}}
    \toprule
    Category & Cond. & ADU or sampled statement & Reference & Prediction/
    assessment \\
    \midrule
    Correct claim
      & B
      & This week shows you acted in line with your values and priorities.
      & \textsc{Claim}
      & AM: \textsc{Claim} \\

    Context resolves error
      & A
      & Movement raises energy and keeps joints flexible.
      & \textsc{Premise}
      & ADU-only: \textsc{Claim}; context: \textsc{Premise} \\

    Correct non-argumentative unit
      & A
      & You knelt by the raised beds, sleeves damp.
      & \textsc{None}
      & AM: \textsc{None} \\

    Rhetorical encouragement
      & B
      & Keep up the great work, as each step you take adds meaning to your
        life and those around you.
      & \textsc{None}
      & AM: \textsc{Claim} \\

    Mixed ADU
      & B
      & You could increase your cycling trips since it combines physical
        exercise with a bit of fun.
      & \textsc{Premise}
      & AM: \textsc{Claim} \\

    Persona-supported statement
      & A
      & You called your children once.
      & \textsc{Grounded}
      & Supported by recorded weekly calls \\

    Plausible but unsupported
      & B
      & Continue engaging in what brings you happiness, and know that each
        effort you make nurtures your well-being.
      & \textsc{Unsupported}
      & No supporting persona information \\
    \bottomrule
  \end{tabularx}
\end{table}

\subsubsection{Persona Grounding and Plausibility}


The examples in Table~\ref{tab:error-examples} illustrate the distinction. ``You called your children once'' was grounded because the corresponding activity and frequency were represented in the persona profile. In contrast, the general encouragement to continue activities that ``nurture your well-being'' was coded as unsupported because the sampled assertion was not supported by the available profile. The statement may appear reasonable as general narrative encouragement, but plausibility is not equivalent to persona grounding.

No direct contradictions were identified in the sampled grounding analysis. Consequently, the present data support an analysis of grounded, unsupported, and non-applicable content, but not a qualitative account of contradiction errors. 
Moreover, the grounding labels were assigned independently by three researchers, labels were determined by majority vote among the three independent annotations, with disagreements resolved through discussion; no automated grounding classifier or statement-level human-acceptance judgement was evaluated. The Phase II perceptions provide broader evidence about narrative relevance and relatability, but they were not linked to these sampled Phase III statements.


\endgroup




\section{Discussion} \label{sec_discussion}

\rev{The findings suggest that perceived purpose in narrative-based dialogue is shaped jointly by the intended communicative structure and the reader’s interpretation.} It cannot be evaluated solely in terms of whether the intended dialogue type is recognised. Participants frequently attributed multiple purposes to the same narrative, suggesting that inquiry, deliberation, and persuasion \rev{may overlap in narrative form and be interpreted differently from their formal structure.}
\rev{The descriptive cross-tabulation also suggests that cultural relatability may matter for future-use intention, although no inferential association or independence from formal argumentation structure was tested. At the same time, 30 of 55 participants indicated that they would not use similar functionality in the future, showing that moderate relevance or cultural relatability did not necessarily translate into willingness to adopt the storytelling functionality. Together, the findings show that formal narrative structure alone does not determine how a purposeful narrative is interpreted or whether similar functionality is considered acceptable.}

\begingroup
\color{black}


The experts' responses also indicate that creativity involves a trade-off rather than a uniformly beneficial design choice. Greater creative expression could make a narrative richer or more engaging, but it could also introduce linguistic complexity, unnecessary elaboration, or speculative content. 
\rev{In Phase II, perceived-creativity ratings were slightly lower for the high-creativity versions, while narrative appreciation did not differ significantly. These findings support treating creativity as a configurable design choice rather than assuming that greater creative expression improves narrative experience.}


The difficulty in distinguishing inquiry from more directive purposes also has implications for applying formal argumentation theory in narrative interaction. Walton and Krabbe distinguish dialogue types according to their communicative goals \cite{walton1995dialogues}, while argumentation schemes provide recurring structures for particular patterns of reasoning \cite{walton2008argumentation}. Our findings suggest that these formal distinctions remain useful for structuring generation, but do not necessarily determine how the resulting narrative is interpreted by a reader. The same argumentative content may function as inquiry when interpreted narrowly as information-seeking, but as deliberation when interpreted in the broader context of considering future action. This contextual dependence is also relevant to argument mining, which seeks to identify argumentative components and relations in text \cite{lippi2016argumentation,mochales2011argumentation}: identifying claims and premises does not by itself establish the dialogue purpose that a reader attributes to them. Thus, formal argumentation provides useful structure for generation and computational inspection, while purpose in person-tailored storytelling should also be evaluated through users’ interpretations and the broader dialogue context.

Finally, the Phase I comments concerning cultural interpretation show that grounding extends beyond factual correspondence with a structured activity profile. A narrative may remain compatible with the supplied information while still appearing culturally unfamiliar or personally inappropriate. Person-tailored storytelling therefore requires attention to both informational grounding and the social and cultural context in which activities acquire meaning, which was taken into consideration in Phase II. 

\endgroup



\rev{The substantial proportion of participants who reported that they would not use similar functionality indicates that moderate perceived relevance alone was insufficient to establish acceptance. Relatively high persona-relatedness ratings across future-use groups further suggest that identifying with a fictional persona was not, by itself, sufficient to distinguish willingness to use the functionality. Future studies will engage older adults over a longer period of time to investigate how the use of their own activities relates to experiences of relevance, cultural relatability, and purpose.}  

\rev{The Phase II questionnaire items were study-specific rather than validated psychometric scales. The findings should therefore be interpreted as descriptive evaluations of participants’ experiences in this study rather than as measurements of validated psychological constructs.}


\rev{The Phase III evaluation provides a complementary technical perspective on these findings. The ablation produced a differentiated result. Removing argumentation-scheme guidance did not result in a statistically detectable reduction in persona grounding (Table~\ref{tab:grounding_conditions}). However, scheme guidance substantially increased explicit premises while claim counts remained essentially unchanged; it also reduced non-argumentative units and increased claim-premise co-presence (Table~\ref{tab:statistical-comparisons}). These outcomes clarify that persona grounding and argumentative support represent different properties of a generated narrative. A claim can be consistent with the persona even when the story does not explicitly state why it follows, whereas scheme-guided generation more often verbalised reasons and justifications within the narrative. The demonstrated contribution of the scheme component in this experiment is therefore stronger for explicit reasoning structure than for increasing the number of persona-supported statements. 
In the A-C comparison, the full structured architecture produced more persona-grounded content than the standard baseline. Because this comparison changed persona information, grounding instructions, dialogue-purpose guidance, and argumentation-scheme guidance together, it supports the architecture as a package rather than isolating any single component. Unsupported statements nevertheless remained in both structured conditions. Taken together, the A-B and A-C findings indicate complementary roles: structured persona information provides the external basis for grounding, whereas argumentation-scheme guidance primarily makes reasons and justifications more explicit within the narrative. }

\rev{The error analysis also qualifies the role of the reflective layer. Narrative context improved the recognition of premises whose function depended on an adjacent recommendation or conclusion. However, rhetorical encouragement and descriptive content were frequently classified as claims, and mixed ADUs exposed limitations in the single-label annotation scheme. The resulting predictions are therefore suitable for assisted inspection, not autonomous verification.}

\rev{Taken together, these findings qualify how formal argumentation can contribute to human-centred AI systems. Prior work has connected computational argumentation with explainability, human-AI collaboration, and broader approaches to responsible AI \cite{demollin_2020_argument,kilic2023argument,Leofante24}. Our results suggest a complementary role in person-tailored storytelling: argumentation schemes can make reasons more explicit and argument mining can expose aspects of argumentative structure for inspection, but neither ensures that users interpret the intended dialogue purpose nor establishes that generated content is grounded in person-specific information. In practice, these mechanisms therefore complement rather than replace user interpretation and explicit grounding assessment. }

\section{Conclusions} \label{sec_conclusion}


This study examined a hybrid person-tailored storytelling architecture integrating argumentation theory, structured user modelling, LLM-based narrative generation, and argument mining, together with how its narratives were experienced by domain experts and older adults. The evaluation further examined persona grounding, the contribution of argumentation-scheme guidance, and the reliability of argument mining as a reflective inspection layer.
\rev{Older adults rated the persona-based narratives as moderately relevant overall, but their responses were mixed: intended-purpose recognition was limited, particularly for persuasion, participants often selected multiple purposes, and many did not intend future use. Cultural relatability showed a descriptive relationship with future-use intention, but the use of predefined personas limits claims about participant personalisation.}


\rev{Relative to the standard LLM baseline, the full architecture showed substantially greater grounding in the supplied persona information within the sampled statements, although unsupported statements remained. The ablation further indicated that scheme guidance primarily increased explicit premises and claim-premise co-presence; it did not produce a statistically detectable grounding-rate difference from the no-scheme condition.}


\rev{The argument-mining layer recovered some explicit claim and premise structure, particularly with narrative context, but weak recognition of non-argumentative content limits autonomous use. Grounding assessment separately identified both profile-supported and unsupported statements. The reflective layer should therefore be treated as an inspection aid rather than a factual verification or hallucination-detection mechanism.}

\rev{Taken together, the study contributes a hybrid approach to person-tailored computational storytelling in which structured activity information and argumentation guide narrative generation, while argument mining provides a complementary inspection layer. The findings show both the value and limitations of these mechanisms and motivate future evaluation with narratives generated from older adults' own activity information in interactive use.}

\begingroup
\color{black}

\section*{Data and Code Availability}

Materials supporting reproducibility of the computational evaluation, including prompts, generation conditions, predefined personas, generated narratives, annotations, model predictions, and analysis scripts, are available at \url{https://github.com/jayalakshmi2k/reflective-storytelling-tist/}.

\endgroup

\begin{acks}
The authors are grateful to the participating domain experts and older adults. 

Research was partially funded by the project "Collaborative Storytelling in Argument-based Micro-dialogues to Improve Health" funded by the KEMPE Foundation (grant number JCSMK22-0158); the project "Digital Companions as Social Actors: Employing Socially Intelligent Systems for Managing Stress and Improving Emotional Wellbeing" funded by Marianne and Marcus Wallenberg Foundation (Dnr MMW 2019.0220); and The Wallenberg AI, Autonomous Systems and Software Program - Humanity and Society (WASP-HS).

\end{acks}

\bibliographystyle{ACM-Reference-Format}
\bibliography{sample-base}









\end{document}